\documentclass[conference]{IEEEtran}
\IEEEoverridecommandlockouts
\usepackage{cite}
\usepackage{amsmath,amssymb,amsfonts}
\usepackage{graphicx}
\usepackage{textcomp}
\usepackage[english,ruled,linesnumbered]{algorithm2e}
\usepackage{lipsum}
\usepackage{array}
\usepackage{subfigure}
\usepackage{color} % comment

\def\BibTeX{{\rm B\kern-.05em{\sc i\kern-.025em b}\kern-.08em
    T\kern-.1667em\lower.7ex\hbox{E}\kern-.125emX}}
\begin{document}

\title{Incremental Semantic Mapping with Unsupervised On-line Learning\\
}

\author{\IEEEauthorblockN{Ygor C. N. Sousa and Hansenclever F. Bassani, Member, IEEE}
\IEEEauthorblockA{\textit{Center of Informatics - CIn}\\
\textit{Federal University of Pernambuco}\\
Recife, PE, Brazil, 50.740-560 \\
Email: \{ycns, hfb\}@cin.ufpe.br}
}

\maketitle

%\thispagestyle{empty}
%\pagestyle{empty}
%%%%%%%%%%% display page numbers %%%%%%%%%%%%%%
%\thispagestyle{plain}
%\pagestyle{plain}
%%%%%%%%%%%%%%%%%%%%%%%%%%%%%%%%%%%%%%%%%%%%%%%%%%%%%%

%%%%%%%%%%%%%%%%%%%%%%%%%%%%%%%%%%%%%%%%%%%%%%%%%%%%%%%%%%%%%%%%%%%%%%%%%%%%%%%%
\begin{abstract}
This paper introduces an incremental semantic mapping approach, with on-line unsupervised learning, based on Self-Organizing Maps (SOM) for robotic agents. The method includes a mapping module, which incrementally creates a topological map of the environment, enriched with objects recognized around each topological node, and a module of places categorization, endowed with an incremental unsupervised learning SOM with on-line training. The proposed approach was tested in experiments with real-world data, in which it demonstrates promising capabilities of incremental acquisition of topological maps enriched with semantic information, and for clustering together similar places based on this information. The approach was also able to continue learning from newly visited environments without degrading the information previously learned. 
\end{abstract}

%%%%%%%%%%%%%%%%%%%%%%%%%%%%%%%%%%%%%%%%%%%%%%%%%%%%%%%%%%%%%%%%%%%%%%%%%%%%%%%%
\section{Introduction}

The idea of semantic mapping is to provide human-centered models of the environment for robots \cite{pronobis2012} that can be used in communication and reasoning. These methods usually receive as input a flux of low-level sensory data, i.e., from LIDARs, cameras, and expand the traditional mapping models (metric or topological map), including higher level semantic concepts that make sense for humans in terms of communication in natural language~\cite{walter2014}. 

Semantic mapping is applied in different tasks involving interaction between humans and robots. Duvallet \cite{duvallet2015} used a semantic map in a wheelchair, allowing it to be controlled by natural language commands as ``go past the kitchen down the hall and then take a right''. Walter et al. \cite{walter2014} extended this idea by incorporating natural language descriptions of places as another source of sensory data. In another study, Walter et al. \cite{walter2015} used a semantic map to allow the understanding of natural language commands given to a forklift.

Current semantic mapping methods usually populate the map with semantic properties organized in a set of predefined types, such as environment size (eg.:large, small), place category (eg.: kitchen, bathroom) and the objects present (eg.: table, chair, TV, sofa). These properties are typically estimated by supervised classifiers from images, low-level sensory data or other extracted semantic properties \cite{walter2014,pronobis2012,kostavelis2015}.

Regarding the topological aspects of environment mapping, most works on the literature apply an incremental approach, creating the map progressively as the robotic agent navigates. The semantic information recognized in the environment is then incorporated into the map. However, the place categorization of the mapped environments is usually done by methods with non-incremental supervised learning \cite{pronobis2012,kostavelis2013,kostavelis2017}. More recently, authors have tried to modify supervised learning methods to add incremental learning \cite{sunderhauf2016} in order to overcome this limitation. 

For robots that are intended to have a long lifetime, that should keep learning as they interact with humans and navigate through different and changing environments, it is necessary to develop appropriate long-term learning methods that can incorporate knowledge incrementally without degrading its performance or requiring retraining \cite{karaoguz2016}.

Therefore, although the traditional non-incremental supervised learning approaches may provide a good performance in a short time or for specific tasks, they may not be adequate for long term learning, since they would require frequent human intervention to update its architecture to specify new categories to be learned and for retraining with new data.

Alternatively, unsupervised learning approaches could minimize the need for such interventions, especially if associated with incremental learning, allowing the incorporation of new knowledge more easily. Additionally, if on-line training would also be associated, the incorporation of new knowledge as more information is available would be possible and in real time, hence, improving robot reasoning and communication capacities with time. 

In this context, this work presents a semantic mapping approach with unsupervised, on-line and incremental learning, which is based on Self-Organizing Maps (SOM) with time-varying structure \cite{araujo2013}. The proposed approach comprises four modules: (i) a Metric SLAM Module that yields the position of the robot; (ii) an Object Recognition Module that recognizes objects in images; (iii) the Semantic Mapping Module (SEMMAP) that creates a topological map of the environment, enriched with the objects recognized around each topological node, the only semantic information to be used to determine the place categories; and (iv) a place categorization module, denominated OLARFDSSOM, that categorizes the place of each topological node, taking as input the semantic information stored on it. 

The first two modules are not the focus of the present work, and any suitable SLAM and object recognition methods can be used. The last two modules learn incrementally and are trained in an unsupervised on-line fashion. Human supervision is required only for labeling linguistically the new place categories already discovered by the module.

The proposed approach was tested with the real world data provided by Pronobis and Caputo~\cite{pronobis2009} and presented promising results. In all experiments, the model was able to create an adequate topological map and to cluster together similar places visited in the explored environments with few errors. The model was also able to learn from new visited environments without degrading the information previously learned.

The following sections of this article are organized as follows: Section~\ref{sec:relatedwork} presents a short review of the related work on semantic mapping. Section~\ref{sec:proposedapproach} presents the proposed approach, which is evaluated in with the experiments presented in Section~\ref{sec:experiments}. Finally, the conclusions are presented in Section~\ref{sec:conclusion}.

\section{Related Work}
\label{sec:relatedwork}

Most semantic mapping methods found in the literature are typically focused on the automatic interpretation of perceptions~\cite{bastianelli2013}, which includes the inference of the places categories present in the environment. The approach presented by Kostavelis and Gasteratos \cite{kostavelis2013} can be used as an example, the authors introduced a semantically annotated topological mapping model that uses a Support Vector Machine (SVM) to infer the place categories. 

Another example is the model introduced by Pronobis and Jensfelt~\cite{pronobis2012}, it uses multiple sensors to recognize different semantic features of the environment. The recognition of objects and environment appearance are done through computer vision techniques, while the sizes and shapes of the rooms are extracted using lasers scanners. The information acquired is classified by SVM models and a probabilistic model of Chain Graph~\cite{lauritzen2002} is used to infer the categories of places. In addition, Bastianelli et al. \cite{bastianelli2013} and Gemignani et al. \cite{gemignani2016} presented models that incrementally add objects pointed by users to multi-layered semantic maps. They, however, do not perform automatic categorization of places.

Unlike the others, Sunderhauf et al. \cite{sunderhauf2016} presented a model that incrementally learns new categories of place, but still, requires supervision in its learning process. In this work, a convolutional neural network is extended by a one-vs-all Random Forest classifier that learns new place categories in a supervised fashion. 

Despite that one of the mentioned methods enable incremental learning at the places categorization step, all of them require some certain level of supervision and do not conduct on-line training. In the literature of unsupervised machine learning, there are methods that are able to incrementally learn data categories. The methods derived from the Self-Organizing Maps with Time-Varying Structure are good candidates. These are a type of neural network in which nodes compete, cooperate and are created incrementally to cluster the input data. Bassani and Araujo \cite{bassani2015} introduced a SOM model of such kind, called LARFDSSOM, that can also deal with high dimensional input data. This family of models inspired the approach presented in this article, which will be detailed in the next section.

\section{Proposed Approach}
\label{sec:proposedapproach}

The proposed approach comprises four modules, which acquire and organize semantic information from the environment. They are:

\begin{itemize}
	\item{\textbf{I - Metric SLAM}: A method of Simultaneous Localization and Mapping of the environment that returns the current position ($x$,$y$) of the robot. This module was not implemented in this work since the position was already provided in the dataset considered.}
	\item{\textbf{II - Object Recognition}: A method that takes as input an image and recognizes a predefined set of objects that may be present on it, then outputs a vector, $\mathbf{r}$, in which each component represents a level of certainty in the [0,1] interval, where zero means that the respective object was not recognized and one indicates that the object was recognized with a high level of certainty. In this work we use a pre-trained model called Inception-v3 \cite{szegedy2015}, available from \cite{tensorflow2015-whitepaper}.}
    \item{\textbf{III - SEMMAP}: A SOM that builds a topological map of the environment, enriched with semantic information, which, in this work, consists solely of information about the objects recognized around each topological node.}
    \item{\textbf{IV - OLARFDSSOM}: An on-line version of the SOM proposed by Bassani and Araujo \cite{bassani2015} that clusters the semantic information stored on the nodes of SEMMAP into categories that aim to represent the types of places visited (eg.: kitchen, corridor, office, etc.).}
\end{itemize}

\begin{figure}[ht]
\vspace{-0.25cm}
\centering
  \includegraphics[width=0.85\columnwidth]{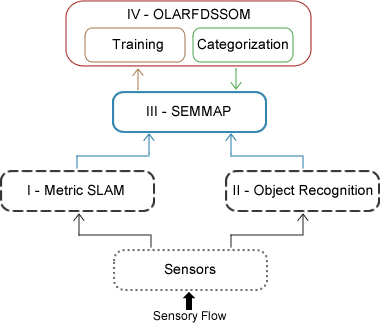}
  \caption{Architecture of the proposed approach. The components with a black dashed outline were not implemented here. The interaction between SEMMAP and OLARFDSSOM indicated in brown represents the training that occurs whenever the agent transits between SEMMAP nodes; and in green, the categorization that can occur at any instant when it is necessary to know the current category of a node in SEMMAP.}
  \label{fig:arquiteturaAbordagem}
\end{figure}

The architecture presented in Fig. \ref{fig:arquiteturaAbordagem} illustrates the information flow between the modules. The sensors that provide data for modules I and II can include LIDARs, Gyroscopes, Regular and Omnidirectional Cameras, etc., according to the techniques employed by modules I and II to determine the position and recognize the basic semantic information from the sensory data. 

The following subsections describe in detail SEMMAP and OLARFDSSOM.

\subsection{SEMMAP}

SEMMAP is a SOM based semantic mapping method that creates topological maps incorporating semantic information captured from the environment. The topological representation starts empty and is incrementally created in a graph form, as the agent moves around. 

The map is represented by a graph $G=(\mathbf{V},\mathbf{E})$, where, $\mathbf{V}=\{v_{j},j=1...k\}$, is a vector of nodes that represents locations on the map, and $\mathbf{E}=\{e_{i},i=1...l\}$, is a vector that represents transition relations between nodes. Each node $j$ on the map is associated with three vectors: $\mathbf{c}_{j}=\{c_{ji},i=1...m\}$, represents the spatial position of the node (center), $\mathbf{o}_{j}=\{o_{ji},i=1...n\}$ ($o_{ji}\in[0,1]$), represents the objects recognized around the position $\mathbf{c}_{j}$, and $\pmb{\phi}_{j}=\{\phi_{ji},i=1...n\}$ is a vector that accumulates the certainty level of the objects recognized, where, $\phi_{ji}\in[0,s_{t}]$ and $s_{t}$ is a parameter that defines an upper limit of accumulation. The vector $\pmb{\phi}_{j}$ is used only to compute $\mathbf{o}_{j}$.

The operation of SEMMAP, as well as of a regular SOM \cite{kohonen1982}, comprises three steps: competition, adaptation, and cooperation. Below we describe how each of these operations is performed in SEMMAP and the Alg.~\ref{alg:mapeamentoSemantico} summarizes them.

\subsubsection{Competition}

In SEMMAP, during the competition step, the nodes on the map compete to cluster the input data, which in the case, is the position received from Module I. The winner of a competition is the most active node according to a radial basis function, i.e., the nearest node to the input position. Whenever the winner node does not achieve a certain activation threshold, a new node is introduced in the map at the position of the input data. 

The position input patterns, $\mathbf{p}$, provided by Module I are presented to the map as the agent moves around the environment, where $\mathbf{p}=(x,y)$, since it usually represents the position of the agent on the horizontal plane. 

When a position input pattern is presented, a competition occurs to determine which node better represents $\mathbf{p}$. The winner of the competition, $s(\mathbf{p})$, is the node that presents the higher activation for the input pattern:
\begin{equation} \label{eq:vencedorCompeticaoSEMMAP}
s(\mathbf{p}) =  \operatorname*{arg\,max}_j [ac(D(\mathbf{p},\mathbf{c}_{j}))].
\end{equation}

The activation of a node, $ac(D(\mathbf{p},\mathbf{c}_{j}))$, is computed as a function of the euclidean distance between the input pattern and the node center:
\begin{equation} \label{eq:calculoAtivacaoSEMMAP}
ac(D(\mathbf{p},\mathbf{c}_{j})) = \dfrac{1}{1 + D(\mathbf{p},\mathbf{c}_{j})},
\end{equation}
where $D(\mathbf{p},\mathbf{c}_{j})$ is calculated as a traditional euclidean distance, as follows:
\begin{equation} \label{eq:distanciaEuclidiana}
D(\mathbf{p},\mathbf{c}_{j}) =  \sqrt{\displaystyle\sum_{i=1}^{m} (p_{i}-c_{ji})^{2}}
\end{equation}

In a competition, if no node achieves the activation threshold $a_{t}$, or if the map is empty, then a new node, $\eta$, is inserted into the map, with $\mathbf{c}_{\eta}=\mathbf{p}$, $\pmb{\phi}_{\eta} = \mathbf{r}$, and $\mathbf{o}_{\eta}$ initialized as per Eq.~\ref{eq:atualizacaoObjetosSEMMAP} (line 9 in Alg.~\ref{alg:mapeamentoSemantico}). Otherwise, the winner node is updated as described following.

\subsubsection{Adaptation}

In the adaptation step, the winner node is adapted to approximate its position to the position of the input data. Therefore, the three vectors associated with the winner node, $s$: $\mathbf{c}_{s}$, $\pmb{\phi}_{s}$ and $\mathbf{o}_{s}$, are updated (lines 14-16 in Alg.~\ref{alg:mapeamentoSemantico}).

The vector $\mathbf{c}_{s}$ is updated taking into account a learning rate $e\in{]0,1[}$ as follows:
\begin{equation} \label{eq:atualizacaoCentrosSEMMAP}
\mathbf{c}_{s}(n+1) =  \mathbf{c}_{s}(n) + e(\mathbf{p}-\mathbf{c}_{s}(n)).
\end{equation}

Although in the current version the semantic properties provided by Module II were not used during the competition, they are accumulated by the winner node. The input patterns representing the objects recognized by Module II, $\mathbf{r}$, are presented to the map as the images are processed, where, $\mathbf{r}=\{r_{i},i=1...n\}$ is a vector containing the certainty level of recognition of each object, $r_{i}\in[0,1]$, and $n$ is the number of different objects that Module II can recognize.

In order to estimate the new value of the object vector, $\mathbf{o}_{s}$, first $\pmb{\phi}_{s}$ is updated. This vector accumulates the evidence about the presence of the objects in the surroundings of the node $s$. This strategy aims to mitigate the problem of object occlusion by collecting data from the different viewpoints. To achieve that, objects certainty values, $\mathbf{r}$, are accumulated in $\pmb{\phi}_{s}$ through a summation limited by $s_{t}$, as follows:
\begin{equation} \label{eq:atualizacaoSomaObjetos}
\phi_{si} =
  \begin{cases}
    s_{t}   & \quad \text{if } \phi_{si} + r_{i} > s_{t} \text{,}\\
    \phi_{si} + r_{i}  & \quad \text{otherwise.}\\
  \end{cases}
\end{equation}

Then, we compute each component, $i$, of the object vector, $\mathbf{o}_{s}$, as a log function of the respective component in $\pmb{\phi}_{s}$:
\begin{equation} \label{eq:atualizacaoObjetosSEMMAP}
o_{si} =  \log_{1 + s_{t}} (1+\phi_{si}),
\end{equation}
where $s_{t}$ is the upper limit used in Eq.~\ref{eq:atualizacaoSomaObjetos}, applied here to ensure that each component $\mathbf{o}_{si}$ is in $[0,1]$ interval.

\subsubsection{Neighborhood and Cooperation}

In the cooperation step, the neighborhood of the winner node is updated. In SEMMAP the neighborhood is formed during the transitions of the agent between two nodes on the map, i.e., nodes that are consecutive winners are connected. The same happens when a new node is inserted into the map: the new node is connected to the previous winner if any (lines 10 and 18 in Alg.~\ref{alg:mapeamentoSemantico}).
  
Differently from the usual SOMs, in SEMMAP the position of the neighbors are not updated and the connections are only used to represent the navigability between the nodes. We intend to better explore the cooperation of the neighborhood in future versions.

Another important aspect of SEMMAP is that the semantic properties stored on the previously visited node are sent to the next module, OLARFDSSOM, whenever a transition occurs, more specifically, when the agent moves from a node $j_a$ to the node $j_b$, the information stored on $\mathbf{o}_{ja}$ is sent to the OLARFDSSOM for training (lines 11 and 18 in Alg.~\ref{alg:mapeamentoSemantico}). This moment was chosen considering that, at that point, node $j_a$ would have accumulated a significant amount of semantic information about its surroundings, that is expected to be sufficient to describe the category of place.
 
\begin{algorithm}[ht]
\small
Initialize parameters $a_{t}$, $e$, $s_{t}$\;
Initialize the map with one node $\eta$ with $\mathbf{c}_{\eta}$ equal to the first position $\mathbf{p}$, $\pmb{\phi}_{\eta}$ equal to the first vector of recognized objects presented $\mathbf{r}$ and $\mathbf{o}_{\eta}$ calculated as per Eq.~\ref{eq:atualizacaoObjetosSEMMAP}\;
Assign $u \leftarrow \eta$, as the last winning node\;
\DontPrintSemicolon
  \SetKwFunction{FMain}{input pattern}
  \SetKwProg{Fn}{foreach}{:}{}
  \Fn{\FMain{$\mathbf{p}$,$\mathbf{r}$}}{
        Present the position $\mathbf{p}$ and the objects $\mathbf{r}$ to the map\;
        Compute the activation of all nodes  (Eq.~\ref{eq:calculoAtivacaoSEMMAP})\;
        Find the winner $s$ with the highest activation ($a_{s}$) (Eq.~\ref{eq:vencedorCompeticaoSEMMAP})\;
        \uIf{$a_{s}<a_{t}$}{
        Create new node $\eta$ and set: $\mathbf{c}_{\eta}\leftarrow \mathbf{p}$, $\pmb{\phi}_{\eta} \leftarrow \mathbf{r}$, $\mathbf{o}_{\eta}$ calculated as per Eq.~\ref{eq:atualizacaoObjetosSEMMAP}\;
        Connect $\eta$ to $u$\;
        Send the vector $\mathbf{o}_{u}$ of the node $u$ to OLARFDSSOM\;
        Assign $u \leftarrow \eta$, as the last winning node\;
        }
        \Else{
        Update the sums vector $\pmb{\phi}_{s}$ of the winner (Eq.~\ref{eq:atualizacaoSomaObjetos})\;
        Update the objects vector $\mathbf{o}_{s}$ of the winner (Eq.~\ref{eq:atualizacaoObjetosSEMMAP})\;
        Update the center vector $\mathbf{c}_{s}$ of the winner (Eq.~\ref{eq:atualizacaoCentrosSEMMAP})\;
        \uIf{$s \neq u$}{
        Connect $s$ to $u$\;
        Send the vector $\mathbf{o}_{u}$ of the node $u$ to OLARFDSSOM\;
        }
        Assign $u \leftarrow s$, as the last winning node\;
        }
  }
\caption{SEMMAP Processing}
\label{alg:mapeamentoSemantico}
\end{algorithm}

\subsection{OLARFDSSOM}

The acronym stands for On-line Local Adaptive Receptive Field Dimension Selective Self-Organizing Map. The proposition here is to introduce an on-line version of LARFDSSOM, a SOM with time-varying structure proposed by Bassani and Araujo \cite{bassani2015}. The LARFDSSOM itself is considered a subspace clustering method, that can find clusters and identify their relevant dimensions, simultaneously, during the self-organization process, with unsupervised and incremental learning. Like the original version, OLARFDSSOM is a general-purpose method that could be applied to different problems, but has so far only been tested in this situation. 

In the present work, the input data to be clustered are the semantic properties (object certainty vectors) collected by SEMMAP, and the clusters formed are expected to represent the different place categories visited by the agent. We consider that LARFDSSOM is a suitable method for this task because it employs a locally weighted distance metric to adjust the relevances of the input dimensions. This is an important property when the input data presents high dimensionality, just as the objects vectors provided by SEMMAP may present. Therefore, LARFDSSOM is able to identify which objects are relevant for determining each place category. As an example, the map can learn that the presence of a TV in a kitchen is not so relevant to recognize this kind of place as the presence of a stove or a sink is. These relevances are automatically adjusted for each cluster.

In OLARFDSSOM the steps of competition, adaptation, and cooperation are done similarly as in LARFDSSOM. We refer the reader to the original paper \cite{bassani2015} for the details about these procedures. However, the original method was not intended to operate with on-line data input, since it is operated in three phases: self-organization, convergence, and clustering. So, in the on-line version presented here, the model is operated in two procedures that occur in parallel: the training and clustering procedures. The main difference between both procedures is that the adaptation and cooperation steps occur only during the training procedure. 

Both procedures are described below and summarized in a form of pseudo-code in Alg.~\ref{alg:autoOrganizacao} (training procedure) and Alg.~\ref{alg:agrupamento} (clustering procedure).

\subsubsection{Training Procedure}

	The training procedure is done after the initialization of the network parameters, whenever a new training pattern $x$ is presented by its inputs. Similarly, as in LARFDSSOM, the first step of the training procedure is the competition, which determines the winner node, $s$. Then, if the activation of the winner node is below the threshold $a_t$, a new node is inserted into the map, at the position of the training pattern. The new node is initialized and connected to other nodes (lines 8-10 in Alg.~\ref{alg:autoOrganizacao}). If the activation of the winner is above or equal to the $a_t$, then, the adaptation and cooperation steps are done (lines 12-13 in Alg.~\ref{alg:autoOrganizacao}). 
    
    In LARFDSSOM, each node $j$ in the map stores a variable, $wins_j$, that accounts for the number of wins of this node with activations not lower than the threshold, since the last reset. A reset occurs after $maxcomp$ competitions, this is the moment when the nodes that present a number of wins below the limit $lp \times maxcomp$ are removed from the map, where $lp$ is a parameter representing the lowest percentage of wins allowed for a node in the map. To avoid the removal of recently created nodes, when a node is created its number of wins is set to $lp\times nwins$, where $nwins$ is the number of competitions that have occurred since the last reset (lines 15-18 in Alg.~\ref{alg:autoOrganizacao}). 
    
Finally, in LARFDSSOM, after the node removal, the number of wins of the remaining nodes is reset to zero. This procedure is not done in OLARFDSSOM to avoid the removal of nodes that represent categories of places that were not recently visited. Therefore, nodes that achieve a number of wins equal or greater than $lp \times maxcomp$ will never be removed from the map.

\subsubsection{Clustering Procedure}

The clustering procedure consists of assigning an input pattern to a cluster. In LARFDSSOM it was done only after the self-organization and convergence phases were finished. In OLARFDSSOM, it may occur at any moment, in parallel with the training procedure, and the result will reflect the current state of the map. 

In OLARFDSSOM each cluster is associated with a unique id that can be retrieved whenever it is necessary to determine which kind of place the agent is in. This id can be further associated with a linguistic label of a place, such as ``kitchen'' or ``office'' for the means of communication in natural language. Thus, the OLARFDSSOM can output the current place category (cluster id) to any SEMMAP node from its semantic properties (object certainty vector) as input pattern, as needed. The Alg.~\ref{alg:agrupamento} details the procedure.

\begin{algorithm}[ht]
\small
Initialize parameters\; 
Initialize the map with one node with $\mathbf{c}_{j}$ initialized at the
first input pattern, $\pmb{\delta}_{j} \leftarrow 0$, $\pmb{\omega}_{j} \leftarrow 1$ and $wins_{j}\leftarrow0$\;
Initialize the variable $nwins\leftarrow1$\; 
\DontPrintSemicolon
  \SetKwFunction{FMain}{input pattern}
  \SetKwProg{Fn}{foreach}{:}{}
  \Fn{\FMain{$\mathbf{x}$}}{
        Present the input pattern $\mathbf{x}$ to the map\;
        Compute the activation of all nodes\;
        Find the winner $s$ with the highest activation ($a_{s}$)\;
        \uIf{$a_{s}<a_{t}$ and $N<N_{max}$}{
        Create new node $j$ and set: $\mathbf{c}_{j}\leftarrow \mathbf{x}$, $\pmb{\delta}_{j} \leftarrow 0$, $\pmb{\omega}_{j} \leftarrow 1$ and $wins_{j}\leftarrow lp \times nwins$\;
        Connect $j$ to the other nodes\;
        }
        \Else{
        Update the vectors of distance $\pmb{\delta}_{s}$, relevance $\pmb{\omega}_{s}$ and center $\mathbf{c}_{s}$ of the winner node and its neighbors\;
        Set $wins_{s} \leftarrow wins_{s}+1$\;
        }
        \If{$nwins\geq maxcomp$}{
        Remove nodes with $wins_{j}<lp \times maxcomp$\;
        Update the connections of the remaining nodes\;
        $nwins \leftarrow 0$\;
        }
        $nwins \leftarrow nwins+1$\;
  }
\caption{OLARFDSSOM Self-Organization}
\label{alg:autoOrganizacao}
\end{algorithm}

\vspace{-0.2cm}

\begin{algorithm}[ht]
\small
\DontPrintSemicolon
  \SetKwFunction{FMain}{input pattern}
  \SetKwProg{Fn}{foreach}{:}{}
  \Fn{\FMain{$x$}}{
        Present the input pattern $\mathbf{x}$ to the map\;
        Compute the activation of all nodes\;
        Find the winner $s$ with the highest activation ($a_{s}$)\;
        Assign $\mathbf{x}$ to the cluster with the id of the winner $s$\;
  }
\caption{OLARFDSSOM Clustering}
\label{alg:agrupamento}
\end{algorithm}

\section{Experiments}
\label{sec:experiments}

In this section, we describe the experiments carried out with the proposed approach. They aimed to evaluate the quality of the obtained maps, both in terms of precision of the topological mapping and of semantic acquisition. In the following subsections, we first describe the dataset considered (Section \ref{sec:dataset}), then the evaluation measures chosen (Section \ref{sec:measures}) and how the parameter adjustment was made (Section \ref{sec:parameters}). The obtained results are presented and discussed in Sections \ref{sec:exptopology} and \ref{sec:expsemantic}.

\subsection{Dataset}
\label{sec:dataset}

In this work, we use the COLD dataset (\textit{COsy Localization Database}) provided by Pronobis and Caputo~\cite{pronobis2009}. The dataset consists of three separated sub-datasets acquired in three different laboratories, each located in a different European city (Freiburg, Ljubljana, and Saarbrucken). Each sub-dataset comprises a sequence of images captured with regular and omnidirectional cameras, along with position data obtained via odometry and laser range scans, as the robotic platform moves in different paths on the facilities. This dataset was chosen to be used in the experiments especially because it contains images and position data.

In total, there are 76 data sequences of 9 different paths in the dataset, 26 of 3 paths in Freiburg, 18 of 2 paths in Ljubljana and 32 of 4 paths in Saarbrucken. However, due to imprecisions found in the position data of several data sequences, only 18 sequences of 6 paths (3 sequences of each path) were used, 6 sequences of 2 paths from Freiburg, and 12 sequences of 4 paths from Saarbrucken. In the 6 used paths, there are 11 different categories of places and each chosen path contains a subset of these categories. 

In this work, we use only the images captured with the regular camera and the respective positions of their acquisition on the environment, $\mathbf{p}=(x,y)$.

For each image of the considered dataset, we run the pre-trained object recognition method, Inception-v3~\cite{szegedy2015}, available on the TensorFlow library~\cite{tensorflow2015-whitepaper}. We defined a set of 18 objects to be recognized. They are: \textit{window shade, bookcase, electric fan, couch, washbasin, soap dispenser, toilet seat, photocopier, monitor, desktop computer, desk, table, chair, banister, microwave oven, stove, dishwasher}, and \textit{toaster}. Each object is recognized with a certainty degree in the [0,1] interval, where zero denotes no object recognized and one the maximum level of certainty. Therefore, each image was transformed into an 18-dimensional vector of certainty levels, $\mathbf{r}$, which was paired with the respective position of acquisition, $\mathbf{p}$.

\subsection{Evaluation Measures}
\label{sec:measures}

In this work, we considered two evaluation measures: Accuracy, which is widely used in the literature for evaluating place categorization, and Clustering Error (CE)~\cite{muller2009}. We consider CE as a better measure for comparing clustering methods that do not necessarily produce a same number of clusters, since it penalizes results with more clusters than necessary, while Accuracy tends to grow with the purity of the clusters, regardless of the number of clusters found.

\subsection{Parameter Adjustment}
\label{sec:parameters}

In order to find adequate values for the several parameters of SEMMAP and OLARFDSSOM, except for the parameters $a_{t}$ and $e$ of the SEMMAP module that were previously manually defined due to the fact that they directly affect the construction of the topological map, we ran a parameter sampling technique known as Latin Hypercube Sampling (LHS) \cite{Helton2005} and recorded the best results achieved by the approach. In the LHS, the parameters are sampled within previously established ranges, where the range of each parameter is divided into subintervals of equal probability and a single value is chosen randomly from each subinterval. The ranges we used for each parameter are presented in Table \ref{table:table-param-ranges} and almost all the experiments described below used the same final parameter configuration, which is also presented in Table \ref{table:table-param-ranges} named as configuration A. The exception was the comparison experiment described in Section \ref{sec:expsemantic} which used another final configuration of parameters in order to best fit the conditions of comparison, the configuration B presented in Table \ref{table:table-param-ranges}.

After analyzing the LHS results, it was possible for us to identify three parameters that affect more significantly the performance of the approach and should be more carefully adjusted: $a_t$, $maxcomp$, and $l_p$ of the OLARFDSSOM. 

\begin{table}[h!t]
\vspace{-0.2cm}
\centering
\caption{Parameter ranges and final configurations.}
\vspace{-0.1cm}
\label{table:table-param-ranges}
\smallskip
\begin{tabular}{lcccc}
\hline
&&\\[-2ex]
\textbf{Parameters} & \textbf{min} & \textbf{max} & \textbf{A} & \textbf{B} \\[0.5ex]
\hline
&&\\[-2ex]
\textbf{OLARFDSSOM} &  &  &  & \\[0.5ex]
\hline
&&\\[-2ex]
Activation threshold ($a_{t}$) & 0.8 & 0.999 & 0.9879 & 0.9668 \\[0.5ex]
Lowest cluster percentage ($l_{p}$) & 0.01 & 0.2 & 0.1914 & 0.1414 \\[0.5ex]
Relevance rate ($\beta$) & 0.001 & 0.1 & 0.0163 & 0.0532  \\[0.5ex]
Max competitions ($maxcomp$) & 5 & 150 & 34 & 89 \\[0.5ex]
Winner learning rate ($e_{b}$) & 0.001 & 0.2 & 0.0118 & 0.0436 \\[0.5ex]
Neighbors learning rate ($e_{n}$) & 0.0001 & $e_{b}$ & 0.0076 & 0.0109 \\[0.5ex]
Relevance smoothness ($s_{}$) & 0.01 & 0.1 & 0.0781 & 0.0453 \\[0.5ex]
Connection threshold ($c_{}$) & 0 & 0.5 & 0.0301 & 0.1108 \\[0.5ex]
\hline
&&\\[-2ex]
\textbf{SEMMAP} &  &  &  & \\[0.5ex]
\hline
&&\\[-2ex]
Activation threshold ($a_{t}$) & - & - & 0.5539 & 0.5539\\[0.5ex]
Learning rate ($e_{}$) & - & - & 0.0139 & 0.0139 \\[0.5ex]
Summation limit ($s_{t}$) & 2 & 15 & 5 & 7 \\[0.5ex]
\hline
\end{tabular}
\end{table}

\subsection{Evaluation of the Topology}
\label{sec:exptopology}

The topology of the maps produced by the SEMMAP was evaluated considering two features: the position of the nodes and the connections between them. The data sequences were presented to the approach and the maps produced by the SEMMAP module were evaluated.

\begin{figure}[ht]
\centering
  \includegraphics[width=0.98\linewidth]{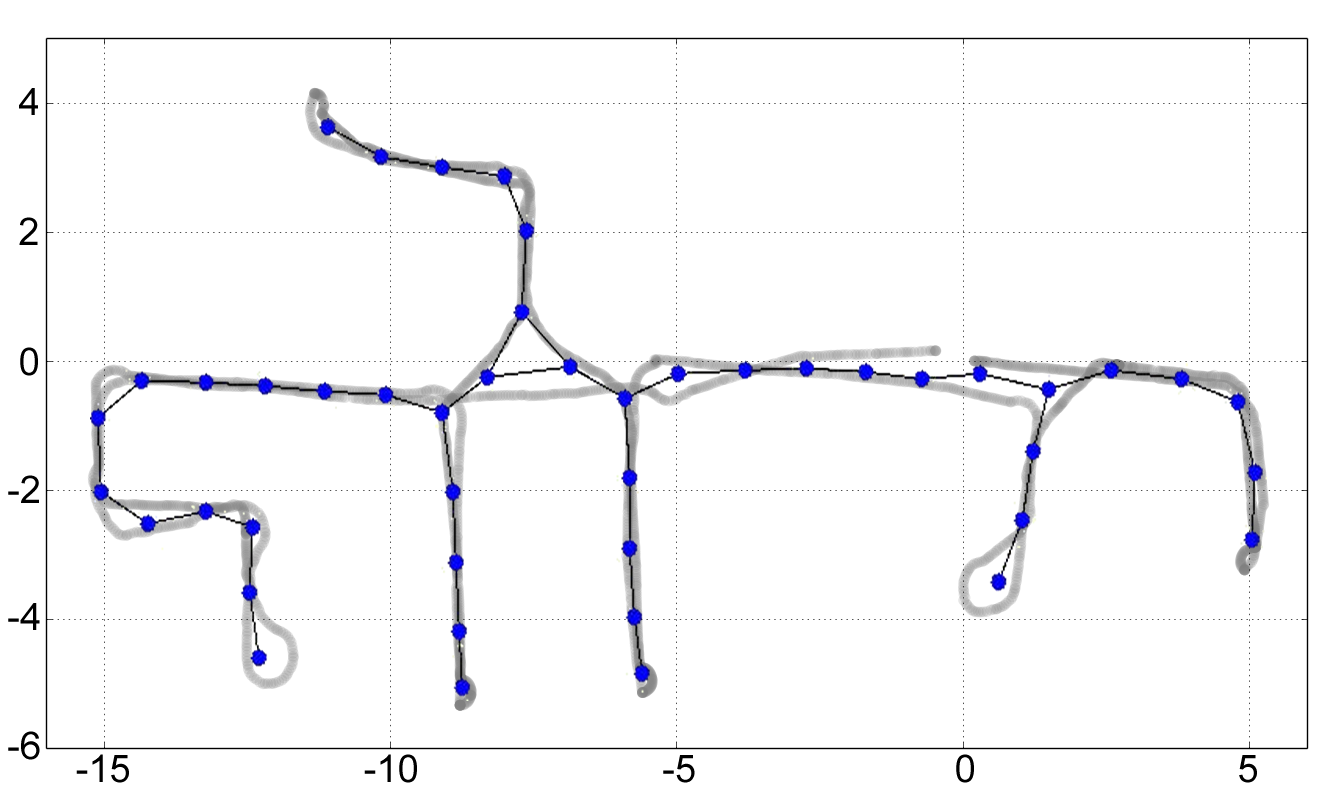}
  \caption{Diagram built from a data sequence from path 2 of the Freiburg sub-dataset. The gray line represents the coordinates provided by SLAM and the blue dots are the nodes created during the topological mapping. The connections between nodes are represented by the black lines.}
  \label{fig:mapa-topologico-freiburg-2-3}
  \vspace{-0.3cm}
\end{figure}

First, the position of each of the 695 nodes created for the 18 maps obtained (coming from the 18 data sequences used) was visually inspected by plotting diagrams in which the node positions are displayed over the positions of the input data. The Fig.~\ref{fig:mapa-topologico-freiburg-2-3} presents a typical example of the results obtained. Such diagrams allowed us to conclude that the nodes were adequately placed in all paths considered.

In order to evaluate the connections between nodes, we evaluated each of the 710 connections formed, verifying if they represent viable paths in the environment. Out of the 710 connections evaluated, only one was incorrectly inserted, what represents an accuracy of 0.9986. We attribute the misplaced connection to a transitory error in the estimated coordinates provided by the dataset.

\subsection{Evaluation of the Semantic Map}
\label{sec:expsemantic}

In the experiments described below, as the semantic map was built by SEMMAP with the input data from Modules I and II, OLARFDSSOM was trained to categorize the places visited, receiving as input the vector of objects stored on the nodes of SEMMAP on each transition. First, we present a comparison of the proposed approach with an image categorization method proposed by Constante et al. \cite{constante2013}. Afterwards, we present an experiment with all the selected paths from the COLD dataset and then an evaluation about the categorization performance over time.

\begin{figure*}[ht]
\centering
  \includegraphics[width=0.73\textwidth]{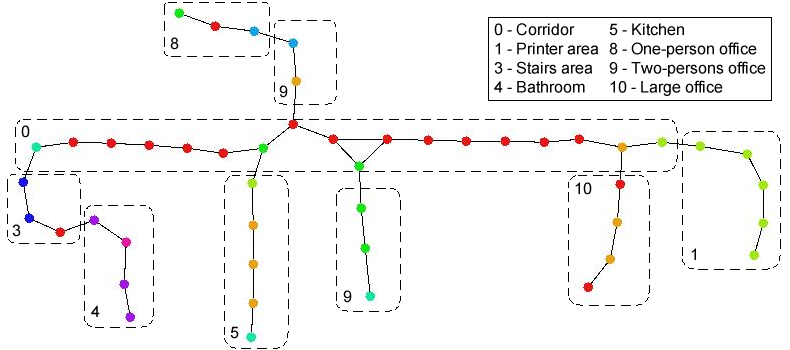}
  \caption{Semantic map of a data sequence from the path 2 of the Freiburg sub-dataset. The colors of the nodes represent the categories found by the model and the dashed squares indicate the expected categories according to the ground truth.}
  \label{fig:mapaSemanticoExperimentoIncremental}
  \vspace{-0.2cm}
\end{figure*}

\subsubsection{Comparison}
A current difficulty in the literature of semantic mapping is the lack of comparability between the results of the different proposed approaches. In order to establish how challenging was the dataset at hand, we considered the results presented by Constante et al. \cite{constante2013} as a reference. This work introduced a place categorization method that uses knowledge from previously labeled places for the categorization of new environments through an unsupervised transfer learning task. The categorization process is done off-line and frame by frame, so only local information contained in each image is analyzed at a time and no semantic map of the environments is created. The method considers two types of image descriptors known as SPMK \cite{lazebnik2006} and SPACT \cite{constante2013}, each of them is tested separately and then combined.

To evaluate the capabilities of the model, the authors used pairs of data sequences from distinct paths from the COLD dataset. The first data sequence was always previously fully labeled and presented to the method that used this information to categorize the second data sequence. The proposed approach was tested on the same data sequences. However, it was trained with unlabeled data of both data sequences and tested only on the second data sequence.  

The results obtained are presented on Tab.~\ref{table:tableresultadoconstante1} in values of Accuracy. As one can notice, the results of the proposed approach are similar to the results presented in \cite{constante2013}, which we considered an satisfactory result, since labeled data is not used in the proposed approach and the method works in an on-line and incremental fashion as it creates the semantic map and recognizes the place categories. It was not possible to carry out a statistical test due to the different nature of both methods.

\begin{table}[ht]
\centering
\caption{Results (Accuracy) of Constante \textit{et al}. \cite{constante2013} compared with the results obtained with the proposed approach (PA).}
\vspace{-0.1cm}
\label{table:tableresultadoconstante1}
\smallskip
\begin{tabular}{p{1.2cm}p{1.5cm}p{0.8cm}p{0.8cm}p{0.8cm}p{0.8cm}}
\hline
&&\\[-2ex]
\textbf{Seq.1/Path} & \textbf{Seq.2/Path} & \textbf{SPMK} & \textbf{SPACT} & \textbf{Both} & \textbf{PA}\\[0.5ex]
\hline
&&\\[-2ex]
Freiburg/1 & Freiburg/2 & 0.7810 & 0.6204 & 0.8267 & 0.6976 \\[0.5ex]
Freiburg/1 & Saarbrücken/1& 0.5019 & 0.5765 & 0.6127 & 0.8148 \\[0.5ex]
Freiburg/1 & Saarbrücken/2& 0.5320 & 0.5620 & 0.6089 & 0.5128 \\[0.5ex]
\hline
\end{tabular}
\end{table}

\subsubsection{Experiment with All Selected Data}

In this experiment, the proposed approach was trained with all previously selected data sequences from both sub-datasets (Freiburg and Saarbrucken), then we evaluated its categorization performance against the ground truth in the data sequences from each sub-dataset separately, conditions: Both/Freiburg and Both/Saarbrucken\footnote{Notation: [training sub-datasets]/[test sub-datasets], ex.: in Both/Freiburg the model was trained with both sub-datasets and tested on Freiburg.}. Additionally, the model was also evaluated in the conditions Freiburg/Freiburg, Saarbrucken/Freiburg, Saarbrucken/Saarbrucken and Freiburg/Saarbrucken. This learning procedure was repeated 30 times with a random selection of data sequences, then we calculated averages and standard deviations for both evaluation metrics. 

The results obtained are presented in Tab.~\ref{table:tableResultadoIncremental1}. As one can notice, in conditions Both/Freiburg and Both/Saarbrucken the results are quite similar or slightly superior to all the others, what was confirmed with a statistical test. This suggests that the model does not degrade in performance as more data is feed into it. Furthermore, it is important to see that the results in conditions Saarbrucken/Freiburg and Freiburg/Saarbrucken are respectively quite similar to Freiburg/Freiburg and Saarbrucken/Saarbrucken, which shows more evidence of the generalizing power of the proposed approach. In Tab.~\ref{table:tableResultadoIncremental1} we display also the number of categories found by OLARFDSSOM in comparison with the ground truth. We notice the method has found a similar number of clusters in all cases. 

An illustration of the typical semantic map obtained with the proposed approach is presented in Fig.~\ref{fig:mapaSemanticoExperimentoIncremental} for a data sequence of a Freiburg path. The color of each node represents one place category found by the model and the dashed squares indicate the expected categories according to the ground truth.

\newcolumntype{C}[1]{>{\centering\arraybackslash}p{#1}}
\begin{table}[ht]
\centering
\caption{Results of the experiment with all selected data. Standard deviations found in parentheses.}
\vspace{-0.1cm}
\label{table:tableResultadoIncremental1}
\smallskip
\begin{tabular}{p{2.6cm}C{1cm}C{1cm}C{0.7cm}C{1.2cm}}
\hline
&&\\[-2ex]
\textbf{Train/Test} & \textbf{CE} & \textbf{Accuracy} & \textbf{Clusters} & \textbf{Categories} \\[0.5ex]
\hline
&&\\[-2ex]
Both/Freiburg & 0.601(0.02) & 0.678(0.02) & 9.40(1.54) & 11 \\[0.5ex]
Freiburg/Freiburg & 0.582(0.02) & 0.650(0.02) & 8.03(1.90) & 8 \\[0.5ex]
Saarbrucken/Freiburg & 0.568(0.03) & 0.660(0.04) & 10.67(2.66) & 9 \\[0.5ex]
Both/Saarbrucken & 0.454(0.02) & 0.540(0.03) & 9.13(1.74) & 11 \\[0.5ex]
Saarbrucken/Saarbrucken & 0.435(0.03) & 0.541(0.02) & 10.46(2.23) & 9 \\[0.5ex]
Freiburg/Saarbrucken & 0.444(0.03) & 0.522(0.02) & 8.13(1.70) & 8 \\[0.5ex]
\hline
\end{tabular}
\vspace{-0.3cm}
\end{table}

\subsubsection{Over Time Evaluation}

\begin{figure*}[ht]
\centering
\subfigure[]{
  \includegraphics[width=0.82\columnwidth]{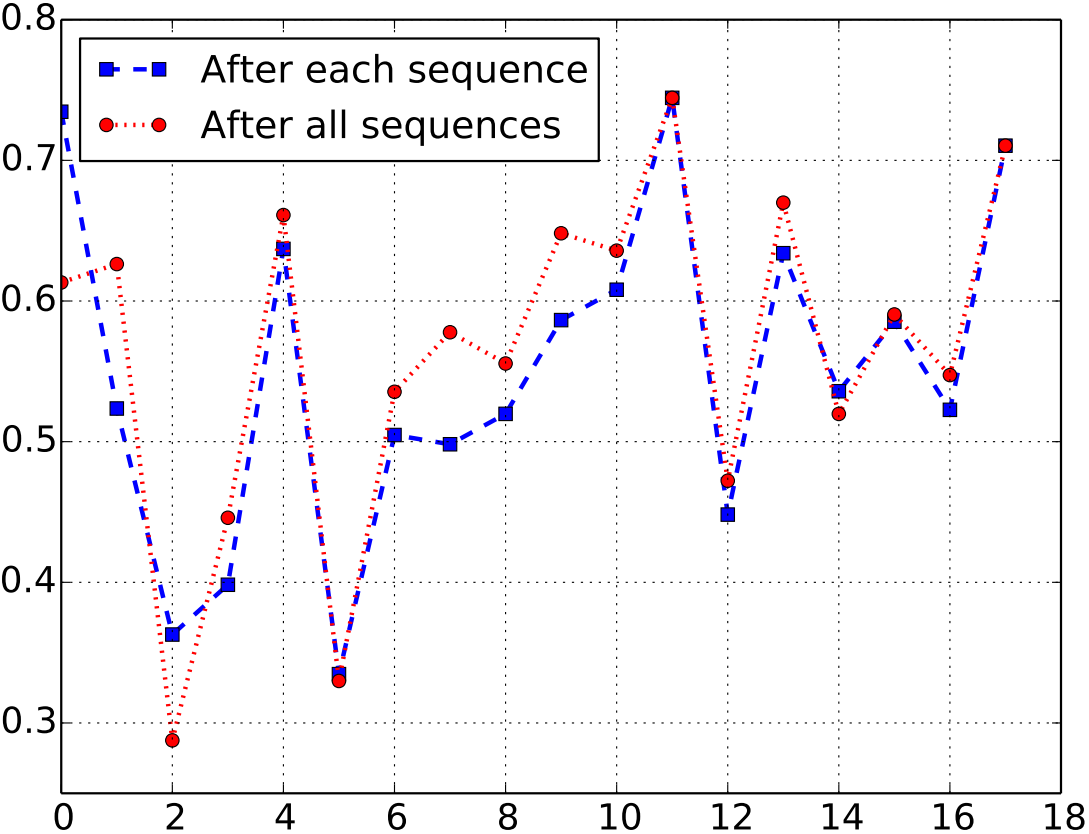}
  \label{subfig:estudodecaso-incremental-1-a}}
\subfigure[]{
  \includegraphics[width=0.82\columnwidth]{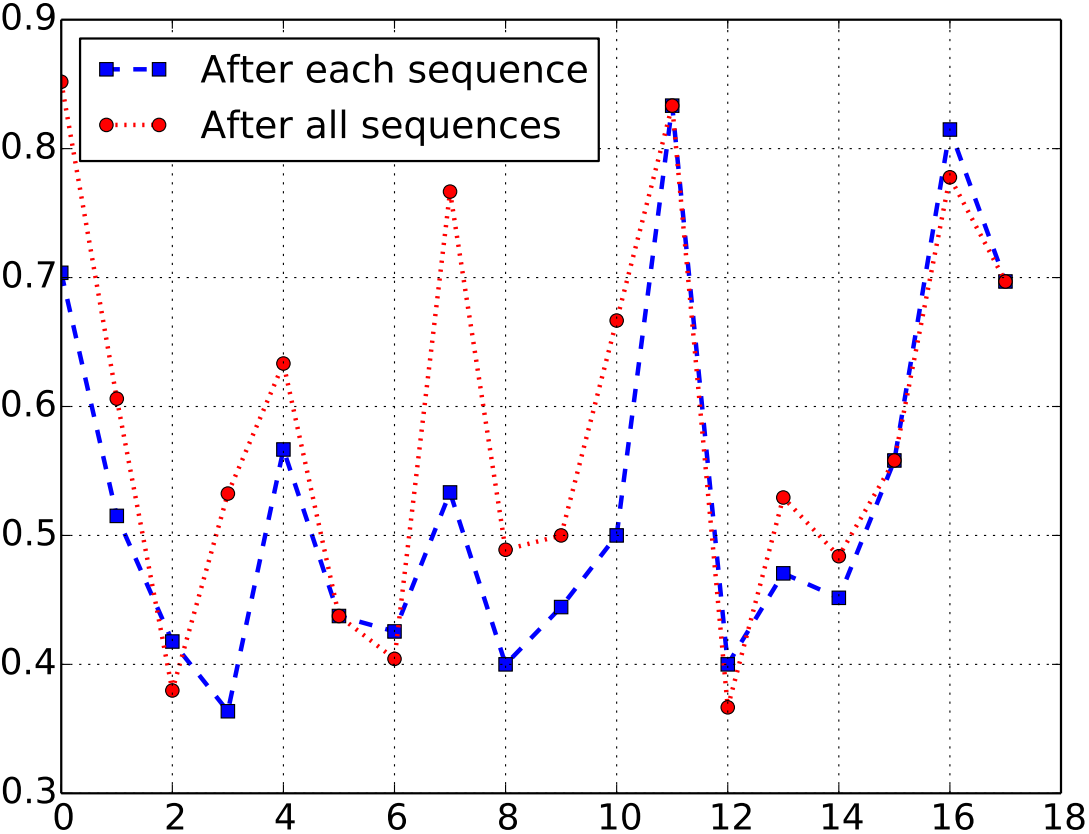}
  \label{subfig:estudodecaso-incremental-1-b}}
\caption{CE (a) and Accuracy (b) obtained (y axis) of each of the 18 selected data sequences (x axis) evaluated in two moments: right after the data sequence training (blue line on the graph) and after training all selected data sequences (red line on the graph).} 
\label{fig:estudodecaso-incremental-1}
\vspace{-0.1cm}
\end{figure*}

In order to evaluate the categorization performance of the proposed approach over time, it was sequentially trained with a random selection of all the data sequences previously selected, with the categorization of each data sequence being evaluated against the ground truth in two moments: right after training the data sequence and after training all data sequences. This aims to verify if the model degrades its performance in the early trained sequences after being trained with other sequences.

The results obtained are graphically shown in Fig. \ref{fig:estudodecaso-incremental-1} and, as can be seen, the values of both evaluation measures after training all sequences were mostly (83,4\% of cases, in CE and 77,8\%, in accuracy) similar or superior to those obtained after training each sequence. This gives us an indication about the behavior of the model, when it is feed with real-time data, displaying its capacity of learning incrementally without apparently degrading its categorization performance in previously learned sequences.

\section{Conclusion}
\label{sec:conclusion}

This paper presented an on-line incremental semantic mapping approach, with unsupervised learning, based on Self-Organizing Maps with Time-Varying Structure. The approach builds topological maps enriched with objects recognized around each node as semantic information. This information is used in real time by an unsupervised learning method to incrementally, and in an on-line fashion, form clusters representing categories of places visited by the agent. The nodes of the topological maps can be categorized at any time with the current state of the unsupervised place category learning method. To the best of our knowledge, this is the first semantic mapping approach with the mentioned characteristics, thus enabling an agent to build semantic maps and learn place categories in real time, as it moves around the environment. 

Moreover, the categorization results obtained were promising, as the place categories found were mostly coherent with the ground truth, grouping together most nodes of the same category, without degrading its capacity with time. The model still presents difficulties for grouping properly nodes located in zones of transition. This was expected since we were using only objects as semantic information and objects can be seen by the camera even before entering a new room.

There are several ways in which the proposed approach could be extended, however, for future work, we first intend to incorporate other kinds of semantic information such place geometry, size, and linguistic data.

\addtolength{\textheight}{-12cm}   % This command serves to balance the column lengths

%%%%%%%%%%%%%%%%%%%%%%%%%%%%%%%%%%%%%%%%%%%%%%%%%%%%%%%%%%%%%%%%%%%%%%%%%%%%%%%%

%%%%%%%%%%%%%%%%%%%%%%%%%%%%%%%%%%%%%%%%%%%%%%%%%%%%%%%%%%%%%%%%%%%%%%%%%%%%%%%%

%%%%%%%%%%%%%%%%%%%%%%%%%%%%%%%%%%%%%%%%%%%%%%%%%%%%%%%%%%%%%%%%%%%%%%%%%%%%%%%%

\section*{Acknowledgment}
The authors would like to thank the Brazilian National Counsel of Technological and Scientific Development (CNPq) for supporting this work.

%%%%%%%%%%%%%%%%%%%%%%%%%%%%%%%%%%%%%%%%%%%%%%%%%%%%%%%%%%%%%%%%%%%%%%%%%%%%%%%%
\bibliographystyle{IEEEtran2007}  
%\bibliography{references}

\end{document}